\documentclass[runningheads]{llncs}
\usepackage{pifont}

\usepackage{amssymb, amsmath, amsfonts}
\usepackage{hyperref}
\usepackage[utf8]{inputenc}
\usepackage{euscript}
\usepackage{mathtools}
\usepackage{subcaption}
\usepackage{graphicx}
\usepackage{caption}
\usepackage{booktabs}
\usepackage{graphicx}
\usepackage{comment}
\usepackage[table]{xcolor}
\graphicspath{Figures/}
\usepackage{multirow}
\usepackage{cleveref}

\usepackage{fix-cm}
\usepackage{threeparttable}
\usepackage{xcolor}
\usepackage{soul}  
\usepackage{hyperref}
\usepackage{mathtools}
\usepackage{booktabs}
\usepackage{multirow}
\usepackage{pifont}
\usepackage{subcaption}
\usepackage{graphicx}
\usepackage[export]{adjustbox}

\begin{document}

\title{CENet: Context Enhancement Network for Medical Image Segmentation}

\titlerunning{CENet: Context Enhancement Network}

\author{Afshin Bozorgpour\inst{1} \and
Sina Ghorbani Kolahi\inst{2} \and
Reza Azad\inst{1} \and
Ilker Hacihaliloglu \inst{3} \and
Dorit Merhof\inst{1,4}}

\authorrunning{\author{A. Bozorgpour et al.}}

\institute{Faculty of Informatics and Data Science, University of Regensburg, Germany \and Department of Industrial and Systems Engineering, Tarbiat Modares University, Tehran, Iran \and Department of Radiology \& Department of Medicine, University of British Columbia, Vancouver, BC, Canada \and Fraunhofer Institute for Digital Medicine MEVIS, Bremen, Germany \\ \email{\{dorit.merhof@ur.de\}}}
%
\maketitle              

\begin{abstract}
Medical image segmentation, particularly in multi-domain scenarios, requires precise preservation of anatomical structures across diverse representations. While deep learning has advanced this field, existing models often struggle with accurate boundary representation, variability in organ morphology, and information loss during downsampling, limiting their accuracy and robustness. To address these challenges, we propose the Context Enhancement Network (CENet), a novel segmentation framework featuring two key innovations. First, the Dual Selective Enhancement Block (DSEB) integrated into skip connections enhances boundary details and improves the detection of smaller organs in a context-aware manner. Second, the Context Feature Attention Module (CFAM) in the decoder employs a multi-scale design to maintain spatial integrity, reduce feature redundancy, and mitigate overly enhanced representations. Extensive evaluations on both radiology and dermoscopic datasets demonstrate that CENet outperforms state-of-the-art (SOTA) methods in multi-organ segmentation and boundary detail preservation, offering a robust and accurate solution for complex medical image analysis tasks. The code is publicly available at \href{https://github.com/xmindflow/cenet}{GitHub}
\end{abstract}


\section{Introduction}
\label{sec:intro}
Medical image segmentation, driven by advancements in deep learning and computer vision, is a critical tool for extracting semantically meaningful information from raw medical datasets. It enables the precise pixel-wise delineation of anatomical structures, organs, and lesions, which are often characterized by diverse shapes, appearances, and pathological conditions~\cite{azad2024medical}. This capability is essential for clinical applications and computer-aided diagnosis systems. Among the most prominent approaches for segmentation are Fully Convolutional Neural Networks (FCNs), particularly the U-Net architecture~\cite{ronneberger2015u} and its variants. These models leverage an encoder-decoder structure to capture multiscale representations: the encoder extracts contextual information, while the decoder upsamples compressed features to produce precise, localized predictions. Skip connections further enhance this process by preserving fine-grained spatial details~\cite{oktay2018attention}, such as boundaries, and enabling the decoder to reconstruct predictions more accurately using high-quality, contextualized features from the encoder.

Despite their strengths, Convolutional Neural Networks (CNNs) inherently struggle to model global contextual relationships due to the limited receptive field of convolutional kernels. This limitation often leads to suboptimal performance in multiscale segmentation tasks involving complex structures~\cite{shamshad2023transformers}. To address this, various strategies have been proposed, including deformable convolutions~\cite{azad2023selfattention}, dilated convolutions~\cite{guo2023visual}, spatial pyramid pooling~\cite{chen2018encoder}, and the integration of attention mechanisms into high-level semantic feature maps~\cite{sohn2022visual}. More recently, the Vision Transformer (ViT)~\cite{dosovitskiy2020image} has emerged as a promising alternative, utilizing self-attention mechanisms to effectively model long-range dependencies and achieve SOTA performance. However, while ViTs excel at capturing global context, they often underperform in modeling local representations and context. Additionally, their quadratic computational complexity makes them inefficient for large-scale applications~\cite{jamil2023comprehensive}. 

Despite advances in CNNs and Transformer networks, their hierarchical structures, reliance on downsampling, and self-attention mechanisms often compromise boundary details and fine-grained semantic representation, limiting their ability to capture multiscale features essential for complex organ and lesion morphologies in medical images. Although hybrid CNN-Transformer architectures~\cite{chen2021transunet,heidari2023hiformer,kolahi2024msa} and localized self-attention mechanisms~\cite{cao2021swinunet,huang2021missformer} have been explored, their focus on global representations and fixed receptive fields restricts accurate segmentation of deformable structures across scales. Many approaches, such as~\cite{wang2022pvt}, focus on body features over edge information, which is crucial for accurate segmentation and detail reconstruction. While studies like~\cite{ma2023lcaunet,10216275} separately integrate fine-grained features (e.g., boundaries), the absence of proper control mechanisms often results in noise, decoder-stage degradation, and inefficient network learning with strong inductive bias.
\\
To address these challenges, we propose the Context Enhancement Network (CENet), a novel framework for medical image segmentation. \ding{202} The Dual Selective Enhancement Block (DSEB) refines fine-grained features by leveraging coarse guidance from the previous decoder, amplifying salient regions while filtering irrelevant information to maintain contextual balance.
\ding{203} The decoder includes the Context Feature Attention Module (CFAM), which uses depth-wise dilated convolutions and a context-aware gating mechanism for multiscale feature representation while addressing over-enhancement through adaptive rectification. \ding{204} Evaluations on radiology (\textit{Synapse}, \textit{ACDC}) and dermoscopic datasets (\textit{PH$^2$}, \textit{HAM10000}) show that CENet outperforms SOTA methods in precise and context-aware segmentation.


\section{Method}
\begin{figure}[!ht]
    \centering
    \includegraphics[width=0.95\linewidth]{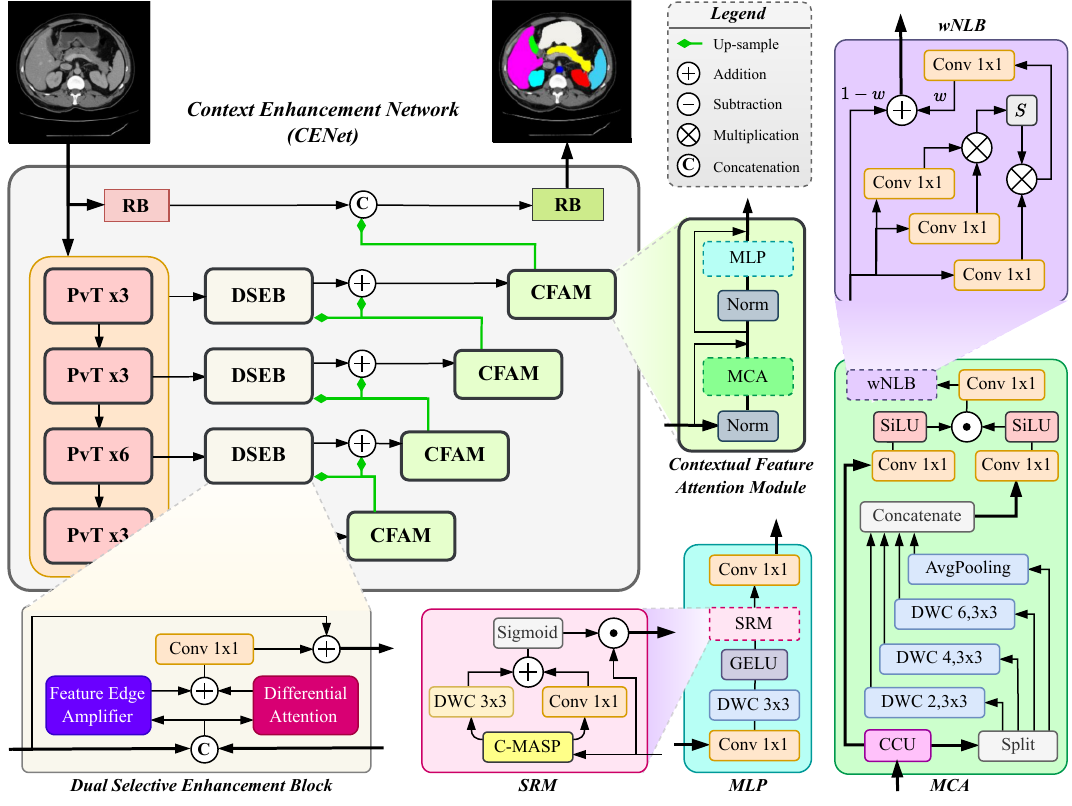}
    \caption{The Contextual Enhancement Network (CENet) uses a pretrained encoder to generate multi-resolution features, processed by the DSEB as skip connections. The decoder refines these features via the CFAM, which includes a CCU, MCA, wNLB, and MLP.}
    \label{fig:net}
\end{figure}

In this study, we propose a novel Context Enhancement Network (CENet) designed to enhance feature representation and improve segmentation accuracy by leveraging hierarchical feature extraction and refined contextual information to effectively capture global contextual dependencies and local spatial details. The network is systematically structured into three key components: a Pyramid Vision Transformer V2 (PvT-V2)~\cite{wang2022pvt} backbone for multi-scale feature extraction, DSEB to contextually enrich skip connections, and CFAM for multi-scale representation and semantic feature refinement. The overall architecture of the proposed method is illustrated in Figure \ref{fig:net}.
\subsection{Dual Selective Enhancement Block (DSEB)}
The DSEB improves CENet's skip connections for precise, boundary-sensitive medical image segmentation. After concatenating encoder features with upsampled CFAM outputs as decoder signals, the input is processed through two stages: the Feature Edge Amplifier (FEA) and Differential Attention (DiffAtt)~\cite{ye2024differential}. The FEA enhances skip connection feature maps by refining edges and spatial details crucial for boundary detection and the localization of small structures. It begins by performing multi-scale downsampling \( \mathcal{D}(F, s) \) and upsampling \( U(F, s) \) on the input feature map \( F \), defined as:  
\begin{equation}
F_{u1} = U(\mathcal{D}(F, s_1), s_0), \quad F_{u2} = U(\mathcal{D}(F, s_2), s_0),
\end{equation}
where \( s_1 = 0.75 \), \( s_2 = 0.5 \), and \( s_0 = 1.0 \) are the scales. The difference between these upsampled features isolates refined edge details, computed as \( F_{\text{edge}} = |F_{u1} - F_{u2}| \). These features are added back to the original feature map \( F \), weighted by \( \lambda \in \mathbb{R}^{d}\), resulting in the enhanced feature map:  
\begin{equation}
\tilde{F} = F + \lambda F_{\text{edge}}.
\end{equation}
The DiffAtt, inspired by NLP, reduces attention noise and selectively enhances meaningful context from the concatenated input features. Query and key vectors are split into two groups, and separate attention maps are computed. By subtracting these maps, issues such as imbalanced token importance in visual contexts (e.g., boundary regions or small structures) are addressed, while redundant attention is removed. The refined attention weights are combined with the value vector, reducing irrelevant context and focusing on critical structures, including fine details like edges. By combining FEA and DiffAtt, DSEB improves feature detail, reinforces localized boundaries while highlighting salient context, surpassing irrelevant regions, and improves segmentation of intricate structures, achieving better overall accuracy.
\subsection{Contextual Feature Attention Module (CFAM)}
The CFAM, a Transformer-based decoder, improves hierarchical feature processing in CENet while avoiding over-enhanced representations. It refines features through four interconnected components for coherent information flow.
First, the \textbf{Channel Calibration Unit (CCU)} processes the input \( X \in \mathbb{R}^{H \times W \times C} \) to recalibrate channel-wise features, enhancing their capacity to capture diverse characteristics for precise segmentation. The CCU employs Global Multi-Aspect Pooling (G-MASP), combining average pooling (\( \mathcal{P}_{\text{avg}}(X) \)), max pooling (\( \mathcal{P}_{\text{max}}(X) \)), and standard deviation pooling (\( \mathcal{P}_{\text{std}}(X) \)) to generate a global descriptor \( \mathbf{g} \in \mathbb{R}^{3C} \).
This descriptor drives a channel-wise attention mechanism that reduces the channel capacity to \( C \) and adaptively reweights feature maps, prioritizing significant channels for improved diversity and representation. The CCU's formulation is given as:
\begin{equation}  
\begin{aligned}  
\hspace{1.5em} \mathbf{g} &= \left[ \mathcal{P}_{\text{avg}}(F); \mathcal{P}_{\text{max}}(F); \mathcal{P}_{\text{std}}(F) \right], \\
\mathbf{s} &= \sigma\left(f_{1\times1}\left(\text{GELU}\left(f_{1\times1}(\mathbf{g})\right)\right)\right), \quad F' = F \odot \mathbf{s},  
\end{aligned}  
\end{equation} 
where $\mathbf{s} \in \mathbb{R}^{C}$ is weights, \( F' \in \mathbb{R}^{H \times W \times C}\) is the reweighted output, \( \odot \) is point-wise multiplication, $f_{1\times1}$ is point-wise convolution, \([;]\) is concatenation, and \( \sigma \) is sigmoid.

Following the CCU, CFAM utilizes the \textbf{Multi-scale Contextual Aggregator (MCA)} to refine \( F' \) by capturing spatial context across multiple scales. The input \( F' \in \mathbb{R}^{C \times H \times W} \) is split into four parts: \( F'_{i} \in \mathbb{R}^{C_{i} \times H\times W} \), where \( F'_1, F'_2, F'_3 \) share equal channel dimensions (\( C_1 = C_2 = C_3 \)), and \( F'_4 \) contains the less than $10$ percent of all the channels, ensuring \( C_1 + C_2 + C_3 + C_4 = C \). Most channels are allocated to the first three branches for the convolution operations, while the fourth branch handles global patterns via average pooling. The splits \( F'_1, F'_2, F'_3 \) are processed with parallel dilated depth-wise convolutions ($f_{dk}$) using dilation rates (e.g., 3, 5, and 8), respectively, while \( F'_4 \) undergoes average pooling. The outputs are concatenated and refined through a \( f_{1 \times 1} \) convolution with SiLU activation. A context-aware gating mechanism then adjusts feature importance, suppressing redundancy and emphasizing salient features, resulting in robust multi-scale representations for improved segmentation. The MCA is expressed as:
\begin{equation}  
\begin{aligned}  
F_{\text{cat}} &= \big[f_{d_3}(F'_1), f_{d_5}(F'_2), f_{d_8}(F'_3), \mathcal{C}_{avg}(F'_4)\big], \\
F_{\text{MCA}} &= \big(\text{SiLU}(f_{1\times1}(F_{\text{cat}})) \odot \text{SiLU}(f_{1\times1}(F'))\big) + F.  
\end{aligned}  
\end{equation}
where $\mathcal{C}_{avg}$ represrnts channel-wise average pooling, \( f_{d_k} \) denotes the depth-wise convolution operator with a kernel size of \( k \times k \) and dilation rate \( d_k \in \{6,8,12\} \).
  
To adjust feature enhancement based on the accumulation of the noise from overly enhanced representations that come from previous layers and corresponding DSEB blocks after feature aggregation, the \textbf{weighted Non-local Block (wNLB)} prevents noise accumulation by modelling long-range spatial dependencies and adaptively denoising features while preserving critical details. It uses a self-attention mechanism with a learnable weighting parameter as a
specific instance of non-local (NL) operations~\cite{wang2018non}. 
The final stage of the CFAM uses an enhanced MLP block, equipped with \textbf{Spatial Calibration Module (SRM)}. SRM applies Channel-wise Multi-Axis Spatial Pooling (C-MASP) using parallel pooling \((\mathcal{C}_{avg}, \mathcal{C}_{max}, \mathcal{C}_{std})\) to create a spatial descriptor \(G \in \mathbb{R}^{3 \times H \times W}\). This descriptor is processed by parallel point-wise convolution that captures pixel-wise relations, and depth-wise convolution (\(f_{k}^{dw}\)) captures neighborhood interactions. The combined output ($S$) recalibrates the feature map via element-wise multiplication, then passes through a linear layer in the enhanced MLP for rich feature representations. Formally:  
\begin{equation}  
\begin{aligned}  
S &= \sigma(f_{1\times1}(G) + f_{k}^{dw}(G)), \quad   
G = \big[\mathcal{C}_{avg}(F_{\text{MCA}}), \mathcal{C}_{max}(F_{\text{MCA}}), \mathcal{C}_{std}(F_{\text{MCA}})\big], \\
&\hspace{2.5cm}F_{\text{recal}} = F_{\text{MCA}} \odot S.  
\end{aligned}  
\end{equation}

\section{Experiments and Results}
\subsection{Datasets and Implementation Details}
The performance of CENet was evaluated on four datasets. The model, developed in PyTorch and trained on an NVIDIA A100 GPU (80GB), used the ImageNet-pretrained PVTv2-b2 encoder~\cite{wang2022pvt} at a 224 × 224 input resolution. For the \textit{Synapse} dataset (30 CT scans), 18 scans were used for training and 12 for validation~\cite{landman2015miccai}, following TransUNet’s protocol~\cite{chen2021transunet}. Training involved 250 epochs, a batch size of 16, and an SGD optimizer with a 0.05 learning rate. On the \textit{ACDC}~\cite{bernard2018deep} dataset (100 cardiac MRI scans), the split was 70 training, 10 validation, and 20 testing cases, with 150 epochs, a batch size of 12, and an Adam optimizer (learning rate 0.0001). For skin lesion segmentation, the \textit{PH2} dataset (80 samples) and \textit{HAM10000} (10,015 images) were trained for 100 epochs with a batch size of 16 and an Adam optimizer (learning rate 0.0001), using preprocessing/augmentation from~\cite{azad2024medical}. The model also integrates BDoU Loss~\cite{sun2023boundary}.


\subsection{Results}
\begin{figure}[!t]
    \centering
    \includegraphics[width=\linewidth]{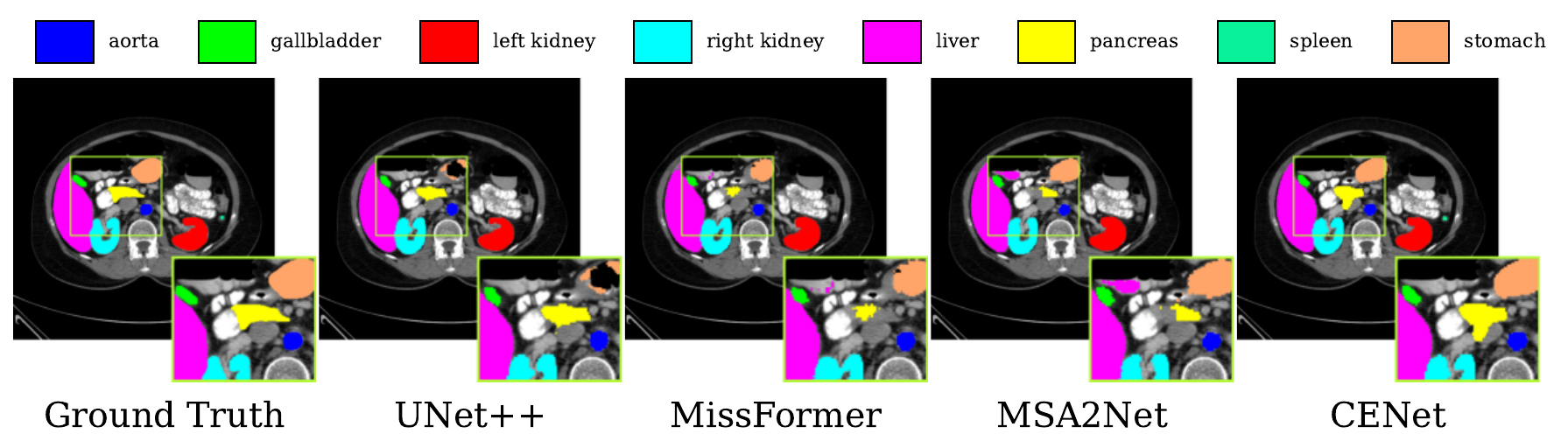}
    \caption{Visual comparison of the proposed method versus others on the Synapse dataset.}
    \label{fig:synapse}
\end{figure}

\textbf{Radiology:} The performance of CENet on radiological datasets was evaluated, with \Cref{tab:synapse} presenting the quantitative results on the Synapse dataset using DSC and HD metrics. Our approach significantly outperforms existing CNN-based methods. CENet also shows enhanced learning capabilities compared to Hybrid models, with improvements of 0.29\% over MSA$^2$Net, respectively. These results underscore CENet's ability in segmenting various organs. \Cref{fig:synapse} provides a visual representation of CENet's performance in segmenting various organs, demonstrating CENet's accuracy in multi-scale segmentation of the kidneys, pancreas, and stomach. Furthermore, \Cref{tab:acdc} emphasizes the effectiveness of our method against SOTA approaches on the ACDC dataset for cardiac segmentation in MRI images. CENet achieves the highest average DICE score of 92.18\%. Moreover, CENet excels in all three organ segmentation tasks, regardless of morphological variations.
\begin{table*}[!t]
    \centering
    \caption{Evaluation results on the Synapse dataset (\textcolor{blue}{blue} indicates the best and \textcolor{red}{red} the second best results).}
    \resizebox{\textwidth}{!}{\begin{tabular}{l|l|l|cccccccc|cc}
    \toprule
     \multirow{2}{*}{Methods}& \multirow{2}{*}{Params}& \multirow{2}{*}{FLOPs}& \multirow{2}{*}{Spl.}&  \multirow{2}{*}{RKid.}& \multirow{2}{*}{LKid.}&  \multirow{2}{*}{Gal.}&  \multirow{2}{*}{Liv.}&  \multirow{2}{*}{Sto.}& \multirow{2}{*}{Aor.}& \multirow{2}{*}{Pan.}& \multicolumn{2}{c}{Average}\\
     \cline{12-13}
     & & & & & & & & & & &DSC$\uparrow$ &HD95$\downarrow$\\

     \midrule\midrule
     R50 U-Net ~\cite{chen2021transunet}& 30.42 M& -&85.87 &78.19 &80.60 &63.66 &93.74 &74.16 &87.74  &56.90 &74.68 &36.87\\


     TransUNet~\cite{chen2021transunet}& 96.07 M& 88.91 G&85.08 &77.02 &81.87 &63.16 &94.08 &75.62 &87.23  &55.86 &77.49 &31.69\\

     
     Swin-UNet~\cite{cao2021swinunet}& 27.17 M& 6.16 G& 90.66& 79.61& 83.28& 66.53& 94.29& 76.60& 85.47& 56.58& 79.13& 21.55\\
     
     
     
     HiFormer-B~\cite{heidari2023hiformer}& 25.51 M& 8.05 G& 90.99& 79.77& 85.23& 65.23& 94.61& 81.08& 86.21& 59.52& 80.39& 14.70\\
     
     

     VM-UNet~\cite{ruan2024vm}& 44.27 M& 6.52 G&  89.51&  82.76& 86.16& 69.41& 94.17& 81.40& 86.40&  58.80& 81.08& 19.21\\



     PVT-EMCAD-B2~\cite{rahman2024emcad} & 26.76 M & 5.60 G & 92.17& 84.10& 88.08& 68.87& 95.26& 83.92& 88.14 & 68.51& 83.63& 15.68\\

     MSA$^{\text{2}}$Net~\cite{kolahi2024msa} & 112.77 M& 15.56 G& \textcolor{red}{92.69}& 84.24& 88.30& \textcolor{blue}{74.35}& \textcolor{red}{95.59}& \textcolor{red}{84.03}& \textcolor{blue}{89.47}& \textcolor{red}{69.30}& \textcolor{red}{84.75}& \textcolor{red}{13.29}\\


     2D D-LKA Net~\cite{azad2023selfattention} & 101.64 M & 19.92 G & 91.22& \textcolor{red}{84.92}& \textcolor{red}{88.38}& \textcolor{red}{73.79}& 94.88& \textcolor{blue}{84.94}& 88.34& 67.71& 84.27& 20.04\\
    
     \midrule
     \textbf{CENet (Ours)}& 33.39 M & 12.76 G & \textcolor{blue}{93.58}& \textcolor{blue}{85.08}& \textcolor{blue}{91.18}& 68.29& \textcolor{blue}{95.92}& 81.68& \textcolor{red}{89.19} & \textcolor{blue}{70.71}& \textcolor{blue}{85.04}& \textcolor{blue}{8.84}\\


     
     \bottomrule
     
     \end{tabular}}\vspace{0.5em}
    \label{tab:synapse}
\end{table*}

\begin{figure*}[!t]
  \centering
  \setlength\tabcolsep{1pt} 
        \begin{tabular}{@{} c!{\vrule width 2pt}c @{}}
            \begin{tabular}{ccc}
                \multicolumn{3}{c}{\textit{PH$^2$}}\\
                \includegraphics[width=0.158\textwidth]{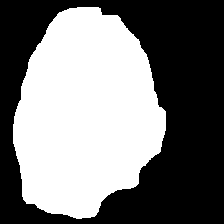} &
                \includegraphics[width=0.158\textwidth]{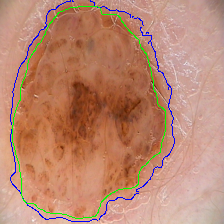} &
                \includegraphics[width=0.158\textwidth]{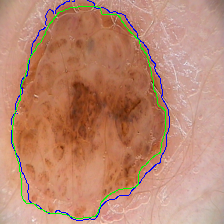} 
            \end{tabular} 
            &
            \begin{tabular}{ccc}
                \multicolumn{3}{c}{\textit{HAM10000}}\\
                \includegraphics[width=0.158\textwidth]{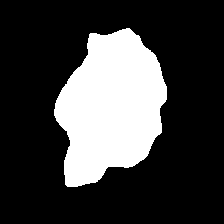} &
                \includegraphics[width=0.158\textwidth]{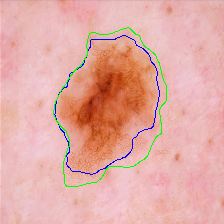} &
                \includegraphics[width=0.158\textwidth]{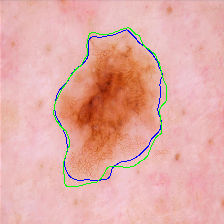} 
            \end{tabular} 
            \\
            \begin{tabular}{ccc}
                \includegraphics[width=0.158\textwidth]{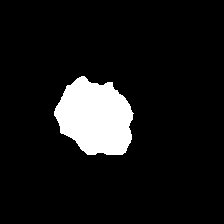} &
                \includegraphics[width=0.158\textwidth]{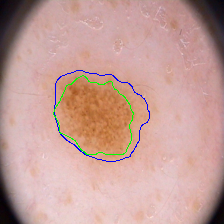} &
                \includegraphics[width=0.158\textwidth]{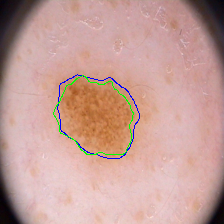} \\
                GT & UCTransNet & CENet
            \end{tabular}             
            &
            \begin{tabular}{ccc}
                \includegraphics[width=0.158\textwidth]{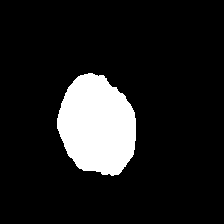} & \includegraphics[width=0.158\textwidth]{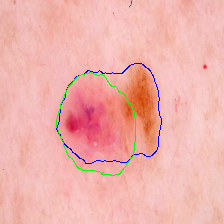} & \includegraphics[width=0.158\textwidth]{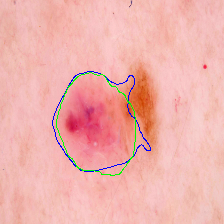} \\
                GT & TransUNet & CENet
            \end{tabular} \\
            
        \end{tabular}
  \caption{Qualitative comparison of CENet and previous methods across skin benchmarks.}
  \label{fig:results-skins}
\end{figure*}

\noindent\textbf{Dermoscopy.} \Cref{tab:skin-benchmarks} evaluates our proposed network on two skin lesion segmentation datasets using DSC and ACC metrics. CENet outperforms CNN-based, Transformer-based, and hybrid methods, demonstrating superior performance and generalization. It surpasses Swin-Unet~\cite{liu2021swin}, U-Net~\cite{ronneberger2015u}, and UCTransNet~\cite{wang2022uctransnet}, achieving DSC score improvements of 0.58\% and 1.45\% on PH$^2$ and HAM10000 datasets, respectively. Moreover, \Cref{fig:results-skins} showcases CENet's superiority in capturing intricate structures and producing precise boundaries through effective boundary integration.

\begin{table}[!ht]
    \centering
    \captionof{table}{Evaluation results on the skin benchmarks (\textit{PH$^2$} and \textit{HAM10000}) and \textit{ACDC} dataset.}
    \resizebox{\textwidth}{!}{
        \begin{tabular}{l p{0.5cm} r}
             \begin{tabular}{l||cc||cc}  
                \hline  
                \multirow{2}{*}{\textbf{Methods}}& \multicolumn{2}{c||}{\textbf{PH$^2$}} & \multicolumn{2}{c}{\textbf{HAM10000}} \\
                \cline{2-5}  
                & \textbf{DSC} & \textbf{ACC} & \textbf{DSC} & \textbf{ACC} \\
                \cline{1-5}  
                U-Net~\cite{ronneberger2015u} & 89.36 & 92.33 & 91.67 & 95.67 \\
                TransUNet~\cite{chen2021transunet} & 88.40 & 92.00 & \textcolor{red}{93.53} & 96.49 \\
                Swin-Unet~\cite{cao2021swinunet} & \textcolor{red}{94.49} & \textcolor{red}{96.78} & 92.63 & 96.16 \\
                DeepLabv3+~\cite{chen2018encoder} & 92.02 & 95.03 & 92.51 & 96.07 \\
                Att-UNet~\cite{oktay2018attention} & 90.03 & 92.76 & 92.68 & 96.10 \\
                UCTransNet~\cite{wang2022uctransnet} & 90.93 & 94.08 & 93.46 & \textcolor{red}{96.84} \\
                MissFormer~\cite{huang2022missformer} & 85.50 & 90.50 & 92.11 & 96.21 \\
                \hline  
                \textbf{CENet} & \textcolor{blue}{95.04} & \textcolor{blue}{97.19} & \textcolor{blue}{94.71} & \textcolor{blue}{97.04} \\
                \hline
            \end{tabular}
            \label{tab:skin-benchmarks}  
             &  & 
             \begin{tabular}{l || c | c c c}  
                \hline  
                \textbf{Methods} & \textbf{aDSC} & \textbf{RV} & \textbf{MYO} & \textbf{LV} \\
                \hline  
                R50+UNet   \cite{chen2021transunet} & 87.55 & 87.10 & 80.63 & 94.92\\
                R50+AttnUNet  \cite{chen2021transunet} & 86.75 & 87.58 & 79.20 & 93.47\\
                TransUNet\cite{chen2021transunet}& 89.71 & 88.86 &  84.53 & 95.73 \\
                ViT+CUP \cite{chen2021transunet}& 81.45 & 81.46 & 70.71 & 92.18 \\
                Swin-UNet \cite{cao2021swinunet} & 90.00 & 88.55 & 85.62 & \textcolor{red}{95.83} \\
                R50+ViT+CUP \cite{chen2021transunet} & 87.57 & 86.07 & 81.88 & 94.75 \\
                MT-UNet \cite{wang2022mixed} & 90.43 & 86.64 & \textcolor{red}{89.04} & 95.62 \\
                MISSFormer~\cite{huang2022missformer} & \textcolor{red}{90.86} & \textcolor{red}{89.55} & 88.04 & 94.99 \\
                \hline  
                \textbf{CENet}  & \textcolor{blue}{92.18} & \textcolor{blue}{90.90} & \textcolor{blue}{89.63} & \textcolor{blue}{95.99} \\
                \hline  
            \end{tabular}
            \label{tab:acdc}
             \\
        \end{tabular}
    }
    \label{tab:skin-acdc}  
\end{table}

\begin{figure}[!htb]
    \centering
    \includegraphics[width=\textwidth]{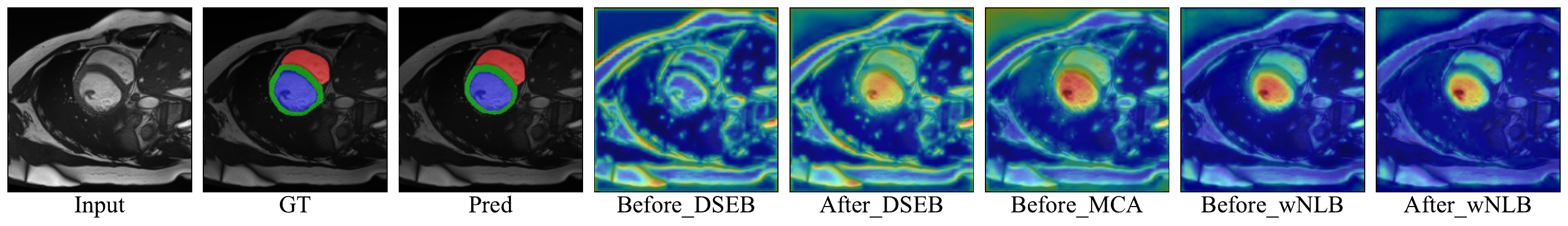}
    \vspace{0.5pt}
    \includegraphics[width = \textwidth]{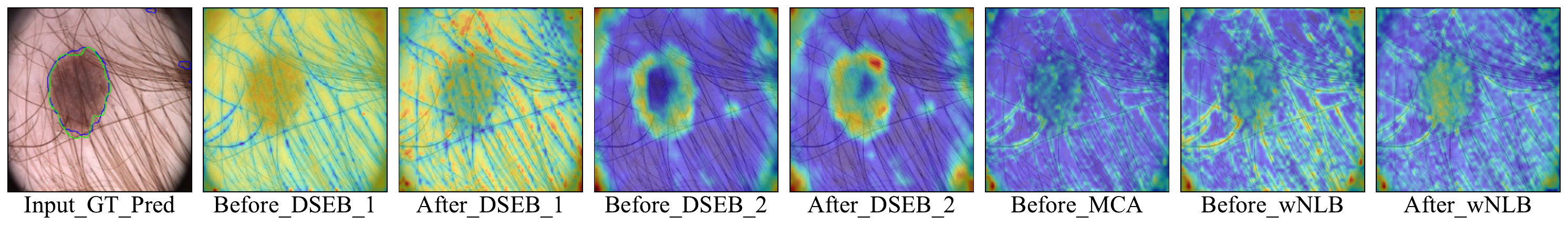}
    \caption{Feature visualization in CENet: first row shows an \textit{ACDC} sample, second row a \textit{PH$^2$} example}
    \label{fig:ablation}
\end{figure}

\noindent\textbf{Ablation Study.} To evaluate CENet’s components, ablation studies were conducted to assess their efficiency, performance, and placement within the DSEB and CFAM modules (\Cref{tab:ablation_components}). Results show that optimal performance occurs when the wNLB is positioned at the end of the MCA and CCU, while the DiffAttn and FEA operate in parallel within the DSEB, outperforming alternative configurations. Feature analyses (\Cref{fig:ablation}) on the \textit{ACDC} and \textit{PH$^2$} datasets further demonstrate that in the first row, attention maps before applying the DSEB are diffuse, failing to focus on key regions. After applying the DSEB, attention becomes more focused, and the wNLB refines MCA multiscale recalibrated feature maps by suppressing irrelevant details and emphasizing salient regions. In the second row, the DSEB enhances feature focus at different CENet levels, while the wNLB acts as a denoising mechanism, suppressing over-attended details (e.g., hairs) and directing attention to critical regions. 
\begin{table}[!ht]
    \centering
    \caption{Effect of different components of CENet on PH$^2$ dataset. \#FLOPs are reported in (G) and P indicates the Parameters in Millions. All results are averaged over three runs. The best results are shown in bold.}
        \begin{tabular}{cc|ccc||c|c||c}
        \toprule
        \multicolumn{5}{c||}{\textbf{Components}} & \multirow{3}{*}{FLOPs (G)} & \multirow{3}{*}{Params (M)} & \multicolumn{1}{c}{\textbf{Performance}} \\
        \cline{1-4}
        \multicolumn{2}{c}{DSEB} & \multicolumn{2}{c}{CFAM} & & & & \multicolumn{1}{c}{\textbf{(DICE)}} \\
        FEA     & \multicolumn{1}{c|}{DiffAtt}     & wNLB & CCU  & & & & PH$^2$ \\
        \midrule
        No   & No     & No   &  No&  & 7.53  & 29.86 & 94.08 \\
        Yes  & No     & Yes  &  No&  & 9.22  & 31.41 & 94.27 \\
        Yes  & Yes    & No   &  No&  & 11.16 & 31.83 & 94.42 \\
        Yes  & Yes    & Yes  &  No&  & 12.84 & 33.37 & 94.79 \\
        Yes  & Yes    & Yes  & Yes&  & 12.84 & 33.39 & \textbf{95.04} \\
        \bottomrule 
        \end{tabular}
    \label{tab:ablation_components}
\end{table}


\section{Conclusion}
The proposed CENet improves context-aware medical image segmentation with a DSEB for boundary-sensitive and context amplification and a CFAM for multiscale representation and reduces feature redundancy. Evaluations demonstrate CENet’s superior accuracy and boundary preservation over SOTA methods, providing a robust solution for complex segmentation tasks.



\bibliographystyle{splncs04}
\bibliography{ref}

\end{document}